\documentclass{article} 
\usepackage[preprint]{colm2026_conference}

\usepackage{microtype}
\usepackage{hyperref}
\usepackage{url}
\usepackage{booktabs}
\usepackage[pdftex]{graphicx}
\usepackage{amsmath}
\usepackage[most]{tcolorbox}
\usepackage{pifont}
\usepackage{multirow}
\newcommand{\cmark}{\ding{51}}
\newcommand{\xmark}{\ding{55}}
\usepackage{subcaption}

\usepackage{lineno}

\definecolor{darkblue}{rgb}{0, 0, 0.5}
\hypersetup{colorlinks=true, citecolor=darkblue, linkcolor=darkblue, urlcolor=darkblue}

\title{Reflection or Re-Generation? Why LLM Revision Fails Where Human Revision Succeeds}

\author{Yefan Tao, \ Gerald Friedland, \ Madhusudhanan Chandrasekaran, \ Luyang Kong \\
Amazon \\
\texttt{\{tayefan, gfriedla, chandmad, luyankon\}@amazon.com}
}

\begin{document}

\ifcolmsubmission
\linenumbers
\fi

\maketitle

\begin{abstract}
Reflection, the ability to revisit and revise prior reasoning, is
central to how humans improve their own answers. Large language
models (LLMs) are increasingly prompted to ``reflect,'' yet whether
this process resembles human revision remains unclear. We introduce the
Human--LLM Reflection Framework (HRF), a controlled two-pass
protocol that compares human and LLM revision behavior under
identical conditions across self-revision, peer-revision, and
cross-agent settings. Using an information-theoretic analysis
grounded in per-iteration cross-entropy reduction, we identify two
distinct failure modes of LLM reflection. On objective reasoning
tasks with finite answer spaces, reflection yields near-zero
information gain ($\Delta I \approx 0$), behaving as neutral
re-generation indistinguishable from independent re-sampling. On
subjective evaluation tasks, reflection produces statistically
significant negative information gain ($\Delta I < 0$), consistently
moving predictions away from the target. Human revision, by contrast,
yields positive information gain across both settings. Cross-agent
experiments localize the failure to the revision step rather than
input quality: LLMs degrade even high-quality human first-pass
responses. Diagnostic analyses---revision conditioned on first-pass
correctness, and oracle-guided revision against a random-reshuffle
baseline---show that which sub-step dominates the failure varies by
task and by model rather than reducing to a single mechanism:
self-error detection is present on objective multiple-choice tasks
but weak on subjective ones, and recovery under an oracle error
signal exceeds a random-reshuffle baseline for some models and falls
below it for others. The unifying account is structural: without
external information, self-conditioned revision cannot reduce
uncertainty about the target, so LLM reflection is better understood
as conditioned re-generation than as genuine error-driven revision.
\end{abstract}

\section{Introduction}
The ability to reconsider and improve prior reasoning is a hallmark of human cognition. In educational psychology, this capacity is studied under the framework of metacognition---the monitoring and regulation of one's own cognitive processes~\cite{dunlosky2008metacognition}. A key insight from this literature is that effective revision involves two separable components: \emph{error detection} (recognizing that something is wrong) and \emph{error correction} (producing a better response)~\cite{nelson1990metamemory, yeung2012metacognition}. This distinction matters because it implies that the ability to generate correct answers does not, by itself, guarantee the ability to identify and fix errors in existing ones.

Large language models (LLMs) have been prompted to engage in analogous behavior. Reflection-based methods instruct a model to revisit a previously generated response and decide whether to retain or revise it~\cite{shinn2023reflexion, ji2023towards, renze2024self}. While some studies report improvements from such prompting, particularly on reasoning tasks, others find that LLMs frequently fail to self-correct and may even degrade their own outputs without external feedback~\cite{huang2023large, li2024hindsight, kamoi2024can}. This tension raises a question that existing evaluations have not resolved: when LLMs are prompted to reflect, are they engaging in genuine error correction, or merely re-generating responses conditioned on their prior output? 

We argue that resolving this question requires two ingredients that prior work has largely lacked. First, a \emph{controlled comparison with human revision under identical conditions}, to establish an empirical reference for what effective reflection looks like. Existing human--LLM comparisons have examined single-pass responses across domains such as moral judgment~\cite{yax2024studying}, theory of mind~\cite{jones2024comparing}, and factual reasoning~\cite{kamoi2024can}, but have not studied the \emph{dynamics} of revision---how answers change from a first attempt to a second. Second, an \emph{information-theoretic characterization of revision dynamics}, to move beyond aggregate accuracy comparisons and ask whether each revision step accumulates task-relevant information or merely reshuffles the model's output distribution.

To this end, we introduce the Human--LLM Reflection Framework (HRF): a controlled two-pass protocol in which both human annotators and LLMs produce an initial response (Pass~1) and then revise it with access to the earlier output (Pass~2). HRF supports three revision settings---self-revision, peer-revision among agents of the same type, and cross-agent revision in which humans revise LLM outputs and vice versa---enabling us to isolate whether failures stem from the revision mechanism itself or from the quality of the input being revised. 

We instantiate HRF on two tasks with discrete, finite answer spaces that enable rigorous information-theoretic analysis: MalAlgoQA~\cite{sonkar2024malalgoqa}, a 4-class mathematical reasoning benchmark, and IMDb-Rating, a 10-class sentiment evaluation task. We complement these with TISER~\cite{bazaga2025learning}, an open-ended temporal reasoning benchmark used for behavioral generalization. We evaluate five representative LLMs spanning multiple families---Llama-3.1-405B,
Claude-3.5-Sonnet, Mistral-Large-0724, GPT-4o-20240806, and
DeepSeek-R1---alongside non-expert human annotators, under identical prompts,
inputs, and evaluation criteria. Two additional models (GPT-5-mini and GPT-5.2) are evaluated in Appendix~\ref{app:additional-models} to assess generalization across model generations.

Our central finding is that LLM reflection exhibits two distinct failure modes, unified by a common information-theoretic explanation. Our contributions are: 
\begin{enumerate} 
\item \textbf{HRF}, a controlled two-pass protocol enabling matched comparison of human and LLM revision behavior across self-revision, peer-revision, and cross-agent settings. 
\item \textbf{A characterization of LLM self-revision as conditioned re-generation.} Under matched conditions, self-revision does not accumulate information about the target: information gain is neutral on objective tasks ($\Delta I \approx 0$) and negative on the subjective task ($\Delta I < 0$), while human revision is positive under identical conditions. We give a unifying information-theoretic account whose central content is a falsifiable prediction: without external information, self-conditioning cannot reduce $H(Y \mid \cdot)$.
\item \textbf{Diagnostic analyses of where revision fails.} Cross-agent matrices localize the failure to the revision step rather than input quality. Revision conditioned on first-pass correctness and oracle-guided revision against a random-reshuffle baseline show that the dominant sub-step---detecting which answers need revision versus producing a better one---varies by task and by model; we report this heterogeneity rather than attributing the failure to a single mechanism.
\end{enumerate}

\begin{figure*}[t]
\centering
\includegraphics[width=0.9\linewidth]{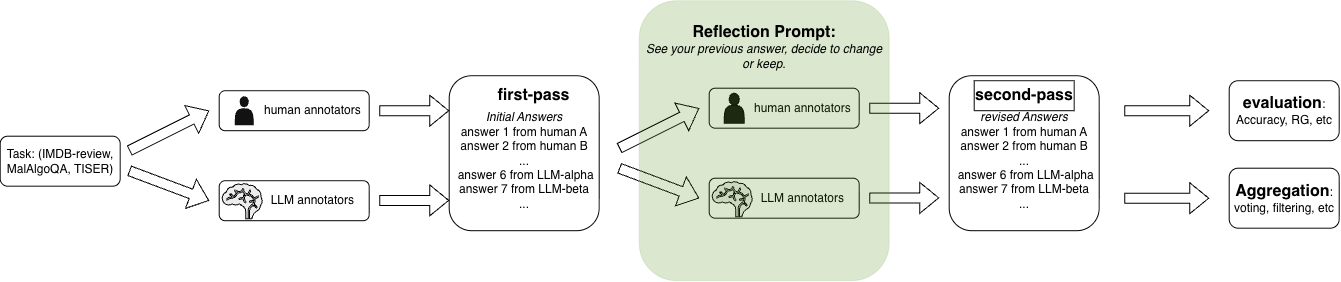}
\caption{
The Human--LLM Reflection Framework (HRF). Human and LLM annotators
independently produce first-pass responses; a reflection prompt then
presents a prior response for optional revision in the second pass.
}
\label{fig:hrf-pipeline}
\end{figure*}

\section{Framework and Experimental Setup}
\label{sec:hrf}

We introduce the Human--LLM Reflection Framework (HRF), a controlled
evaluation protocol for studying reflective revision in humans and
LLMs under identical conditions. The overall pipeline is illustrated
in Figure~\ref{fig:hrf-pipeline}.

\subsection{Two-Pass Protocol and Revision Settings}

HRF follows a two-pass design. In the first pass, an
annotator---human or LLM---produces an initial response based solely
on the task input. In the second pass, the same task is presented
together with a previously generated first-pass response, and the
annotator decides whether to retain or revise it. No additional
information beyond the original input and the supplied prior response
is provided, ensuring that any performance change is attributable to
the revision process itself.

Based on the source of the prior response, HRF distinguishes three
settings. In \emph{self-revision}, the annotator revises their own
first-pass output. In \emph{peer-revision}, the prior response is
produced by a different annotator of the same type, with first-pass
responses randomly reassigned. In \emph{cross-agent revision},
humans revise LLM outputs and vice versa. The cross-agent setting
serves a diagnostic purpose (Section~\ref{sec:reviser-not-source}):
if reflection failures stem from poor first-pass quality, providing
a higher-quality source should help; if the revision mechanism itself
is deficient, performance should degrade regardless of source.

\subsection{Tasks and Datasets}

We evaluate three tasks; the first two have discrete, finite answer
spaces enabling the information-theoretic analysis in
Section~\ref{sec:info-framework}.
\vspace{-3mm}
\paragraph{IMDb-Rating.} A sentiment evaluation task pairing movie
reviews with ground-truth ratings on a discrete 1--10 scale
(908 items).
\vspace{-3mm}
\paragraph{MalAlgoQA \citep{sonkar2024malalgoqa}.} A 4-choice
mathematical reasoning benchmark (709 questions, shuffled options).
\vspace{-3mm}
\paragraph{TISER \citep{bazaga2025learning}.} A temporal reasoning
benchmark requiring consistency under evolving constraints
(250 samples); because responses are free-form, we use TISER for
behavioral analysis only, with correctness assessed by a fixed LLM
judge~\citep{alexandru2025atla}.

All datasets were released after the training-data cutoff dates of
the models evaluated, minimizing contamination risk.

\subsection{Annotators and Evaluation}

\paragraph{LLMs.}
We evaluate five models: Llama-3.1-405B, Claude-3.5-Sonnet,
Mistral-Large-0724, GPT-4o-20240806, and DeepSeek-R1. All use
identical prompts and strict JSON output constraints; sampling
parameters are fixed across conditions. Two additional models
(GPT-5-mini and GPT-5.2) are evaluated in
Appendix~\ref{app:additional-models}.
\vspace{-3mm}
\paragraph{Humans.}
Annotations are collected via Amazon SageMaker Ground Truth,
targeting 5 annotators per item. Inter-annotator agreement
(Krippendorff's $\alpha$) is reported in
Appendix~\ref{app:annotation-details} (Table~\ref{tab:iaa}).

\vspace{-3mm}
\paragraph{Metrics.}
IMDb-Rating uses MSE (lower is better); MalAlgoQA and TISER use
accuracy (higher is better). Reflection gain (RG) quantifies the
relative change from first to second pass:
\begin{equation}
\text{RG} = \frac{\phi(\hat{y}^{(2)}) - \phi(\hat{y}^{(1)})}
            {|\phi(\hat{y}^{(1)})|} \times 100\%,
\end{equation}
where $\phi$ is the task metric, defined so that higher values
indicate better performance. For group-level analysis, we aggregate
via majority voting.

\section{Empirical Overview}
\label{sec:empirical}

\subsection{Reflection Gain Across Tasks}
Table~\ref{tab:main} reports first-pass performance, second-pass performance, and reflection gain for all models and human annotators across the three tasks. Three patterns emerge from this table.

\paragraph{Humans consistently improve; LLMs do not.} Across all three tasks, human annotators exhibit positive reflection gains: $+17.8\%$ on IMDb-Rating, $+6.0\%$ on MalAlgoQA, and $+4.7\%$ on TISER (averages across annotators). In contrast, no LLM achieves consistent positive gains across tasks. Aggregating multiple LLM responses via majority voting partially stabilizes outcomes but does not close the gap to human performance.

\paragraph{Reflection is actively harmful on subjective tasks.} On IMDb-Rating, \emph{every} LLM exhibits negative reflection gain, with degradation ranging from $-5.1\%$ (Mistral) to $-29.2\%$ (Claude). On objective tasks (MalAlgoQA, TISER), reflection effects are smaller in magnitude and directionally mixed, with most models showing near-zero or slightly negative gains. The qualitative difference between subjective and objective tasks is striking: IMDb-Rating is the only setting in which reflection \emph{systematically} degrades all models, suggesting a failure mode distinct from the neutral re-generation observed on objective tasks.

\begin{table*}[t]
\centering
\small
\setlength{\tabcolsep}{4pt}
\begin{tabular}{lccc ccc ccc}
\toprule
& \multicolumn{3}{c}{IMDb-Rating $\downarrow$}
& \multicolumn{3}{c}{MalAlgoQA $\uparrow$}
& \multicolumn{3}{c}{TISER $\uparrow$} \\
\cmidrule(lr){2-4} \cmidrule(lr){5-7} \cmidrule(lr){8-10}
Model & 1st & 2nd & RG
      & 1st & 2nd & RG
      & 1st & 2nd & RG \\
\midrule
Mistral-Large   & 1.764 & 1.855 & \textcolor{red}{$-$5.1\%}
                & 0.750 & 0.743 & \textcolor{red}{$-$1.0\%}
                & 0.456 & 0.464 & \textcolor{teal}{+1.8\%} \\
Llama-3.1-405B  & 2.011 & 2.242 & \textcolor{red}{$-$11.5\%}
                & 0.776 & 0.775 & \textcolor{gray}{$0.0\%$}
                & 0.492 & 0.484 & \textcolor{red}{$-$1.6\%} \\
Claude-3.5      & 1.594 & 2.058 & \textcolor{red}{$-$29.2\%}
                & 0.829 & 0.815 & \textcolor{red}{$-$1.7\%}
                & 0.536 & 0.548 & \textcolor{teal}{+2.2\%} \\
GPT-4o          & 1.760 & 2.021 & \textcolor{red}{$-$14.8\%}
                & 0.801 & 0.787 & \textcolor{red}{$-$1.8\%}
                & 0.528 & 0.528 & \textcolor{gray}{$0.0\%$} \\
DeepSeek-R1     & 1.786 & 1.937 & \textcolor{red}{$-$8.4\%}
                & 0.948 & 0.953 & \textcolor{teal}{+0.6\%}
                & 0.536 & 0.520 & \textcolor{red}{$-$3.0\%} \\
LLM-voted       & 1.718 & 1.844 & \textcolor{red}{$-$7.3\%}
                & 0.851 & 0.889 & \textcolor{teal}{+4.5\%}
                & 0.516 & 0.536 & \textcolor{teal}{+3.9\%} \\
\midrule
Human (avg)     & 2.431 & 1.997 & \textcolor{teal}{+17.8\%}
                & 0.830 & 0.880 & \textcolor{teal}{+6.0\%}
                & 0.382 & 0.400 & \textcolor{teal}{+4.7\%} \\
Human (top-3)   & 1.602 & 1.205 & \textcolor{teal}{+24.8\%}
                & 0.901 & 0.920 & \textcolor{teal}{+2.2\%}
                & 0.490 & 0.513 & \textcolor{teal}{+4.5\%} \\
Human-voted     & 1.750 & 1.510 & \textcolor{teal}{+13.7\%}
                & 0.903 & 0.976 & \textcolor{teal}{+8.1\%}
                & 0.448 & 0.580 & \textcolor{teal}{+29.5\%} \\
\bottomrule
\end{tabular}
\caption{Reflection performance across tasks and passes. Reflection gain (RG) denotes relative improvement from the first to the second pass. For IMDb-Rating the metric is MSE (lower is better); for MalAlgoQA and TISER it is accuracy (higher is better).}
\label{tab:main}
\end{table*}

\paragraph{Model strength does not resolve the problem.} DeepSeek-R1, the strongest model in our evaluation (MalAlgoQA first-pass accuracy $0.948$), achieves only $+0.6\%$ reflection gain on MalAlgoQA and $-3.0\%$ on TISER. The failure of reflection is not confined to weaker models: even the highest-performing LLM cannot reliably convert reflection into improvement. Additional experiments with more recent models (GPT-5 series) confirm this pattern (Appendix~\ref{app:additional-models}).

\subsection{Revision Frequency and Precision}
\label{sec:revision-precision} 
Beyond aggregate accuracy, we examine how often annotators revise and how often those revisions are beneficial. Detailed self- and peer-revision breakdowns are reported in Appendix~\ref{app:self-peer-plots}.
\paragraph{Humans revise less but more precisely.} Humans initiate revisions less frequently than LLMs, but a substantially larger fraction of human revisions are beneficial. In contrast, LLMs revise more frequently, yet beneficial and harmful revisions occur at comparable rates, so increased revision frequency does not translate into effective reflection.
\paragraph{LLMs show stronger self--peer asymmetry.} Both humans and LLMs revise peer responses more frequently than their own, but this asymmetry is substantially larger for LLMs. For LLMs, peer-revision produces larger but less consistent changes, whereas human revisions remain comparatively stable across both settings.

These patterns establish that LLM reflection fails, but do not explain why it fails, nor why the failure differs across task types. The remainder of the paper addresses these questions: Section~\ref{sec:info-framework} introduces the information-theoretic framework, Section~\ref{sec:two-failures-main} identifies two distinct failure modes, and Section~\ref{sec:reviser-not-source} diagnoses their cause.

\section{Information-Theoretic Framework}
\label{sec:info-framework}

To explain why LLM reflection fails—and why it fails differently across
task types—we model reflection as an information acquisition process and
ask whether each revision step reduces uncertainty about the target
answer.

Let $x$ denote the task input, $y^{*}$ the target answer, and
$\hat{y}^{(1)}$ the initial response. A reflection step produces a
revised response by conditioning on the prior answer:
\begin{equation}
\hat{y}^{(2)} \sim p_{\theta}(y \mid x, \hat{y}^{(1)}).
\end{equation}
From an information-theoretic perspective, reflection is effective only
if this conditioning reduces uncertainty about $y^{*}$. We formalize
this via the per-iteration information gain at revision step $k$:
\begin{equation}
\Delta I^{(k)} = I(y^{*};\, \hat{y}^{(k)} \mid x,\, \hat{y}^{(k-1)}),
\end{equation}
where positive values indicate that revision brings the output closer
to the target and negative values indicate that it moves further away.
Effective reflection corresponds to sustained positive $\Delta I^{(k)}$
across iterations.

Because $y^{*}$ is not directly observable, we estimate $\Delta I^{(k)}$
using cross-entropy reduction. For each revision step $k$, we compute
the cross-entropy $H(y^{*} \mid p^{(k)})$ between the model's predicted
answer distribution $p^{(k)}$ and the one-hot ground-truth label
$y^{*}$, where $p^{(k)}$ is estimated from $K = 10$ independent runs
under identical decoding settings. The empirical information gain is
then
\begin{equation}
\Delta I^{(k)}_{\mathrm{emp}} = H(y^{*} \mid p^{(k-1)})
  - H(y^{*} \mid p^{(k)}).
\end{equation}

We are explicit about what this estimator does and does not provide.
With a point-mass target, $H(y^{*}\mid p^{(k)})$ reduces to
$-\log_2 \hat{p}^{(k)}(y^{*})$, so on finite-label tasks
$\Delta I^{(k)}_{\mathrm{emp}}$ is a monotone reframing of the change
in the empirical probability of the correct answer, and on the ordinal
rating task the distance-aware analogue $\tfrac{1}{2}\log_2(\mathrm{MSE}_1/\mathrm{MSE}_2)$
is one-to-one with the MSE ratio. It is therefore not an independent
source of evidence beyond accuracy and MSE, and we refer to it as
\emph{predictive cross-entropy reduction} rather than claiming a new
measure. Its role is twofold: it places tasks scored by accuracy and by
MSE on one bits axis so the two failure modes can be compared, and it
yields a falsifiable prediction---without external information,
self-conditioning cannot reduce $H(Y\mid\cdot)$---that the self-, peer-,
cross-agent, oracle, and human experiments are designed to test.

We apply this analysis to the two tasks with discrete answer
spaces—MalAlgoQA (4-class) and IMDb-Rating (10-class)—using matched
items throughout. TISER is excluded because its free-form answers lack
a finite label space, making entropy-based measures ill-defined; we
rely on behavioral metrics for that task.

\section{Failure Modes and Diagnosis of LLM Reflection}
\label{sec:failure-diagnosis}

\subsection{Two Failure Modes}
\label{sec:two-failures-main}

\begin{figure*}[t]
\centering
\includegraphics[width=0.9\linewidth]{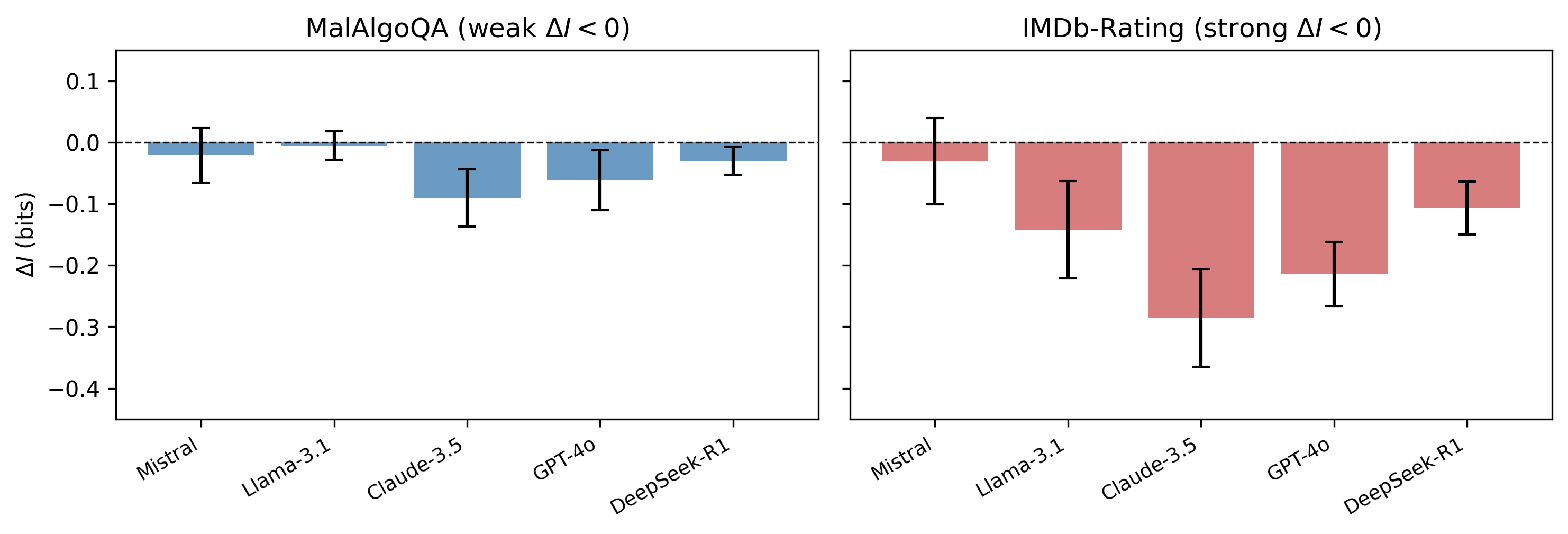}
\caption{Per-model information gain $\Delta I$ on MalAlgoQA (left,
4-class) and IMDb-Rating (right, 10-class). All bars are at or below
zero. On MalAlgoQA, $\Delta I$ is weakly negative and largely
indistinguishable from zero (neutral re-generation). On IMDb-Rating,
$\Delta I$ is substantially negative across all models (distributional
flattening). Error bars show 95\% confidence intervals over $K=10$
runs.}
\label{fig:delta-i-bar}
\end{figure*}

Figure~\ref{fig:delta-i-bar} reports per-model information gain
$\Delta I^{(1)}_{\mathrm{emp}}$ on the two tasks with discrete answer
spaces, estimated via cross-entropy reduction over $K=10$ repeated
runs.

Two distinct patterns emerge:

\paragraph{Objective tasks: neutral re-generation
($\Delta I \approx 0$).}
On MalAlgoQA, all five models produce $\Delta I$ values that are
weakly negative and close to zero. Confidence intervals either include
zero or sit just below it. Repeated-run paired $t$-tests ($K=10$)
confirm that self-revision does not yield statistically significant
improvement over stochastic re-generation for four of five models; the
one exception (Claude-3.5) shows a small but significant
\emph{decrease} (Appendix~\ref{app:stat-tests}). On TISER, the same
statistical analysis reveals a similarly stable pattern: no model
achieves significant positive gain
(Appendix~\ref{app:stat-tests}). Taken together, these results
indicate that on objective tasks, conditioning on the first-pass
response introduces no new information about $y^{*}$---revision is
statistically indistinguishable from independent re-sampling.

\paragraph{Subjective tasks: distributional flattening
($\Delta I < 0$).}
On IMDb-Rating, the picture is qualitatively different. All five
models exhibit negative $\Delta I$, and the effect is substantially
larger in magnitude than on MalAlgoQA. Paired $t$-tests over repeated
runs confirm that all five models show statistically significant MSE
increases under reflection compared to re-generation
(Appendix~\ref{app:stat-tests}). Unlike the near-neutral pattern on
objective tasks, reflection on IMDb-Rating moves predictions away from
the target: second-pass predictions are systematically \emph{further}
from the ground-truth rating than first-pass predictions. We refer to
this pattern as distributional flattening, and Appendix~\ref{app:drift}
provides geometric evidence consistent with predictions drifting toward
a model-internal prior; we note that this is one mechanism consistent
with the data (others, such as variance reduction under conditioning or
anchoring on the prior text, are not ruled out), whereas the negative
$\Delta I$ itself is a direct measurement.

\paragraph{Interpretation.}
Both failure modes stem from the same limitation: without an external
error signal, conditioning on the first-pass response cannot introduce
new information about $y^{*}$. On objective tasks with a small answer
space, this produces neutral re-sampling ($\Delta I \approx 0$). On
subjective tasks, where the model's prior concentrates on a narrower
range than the ground truth, uninformative conditioning triggers
regression toward default predictions, overwriting first-pass outputs
that had captured informative distributional features
($\Delta I < 0$). Geometric analysis in embedding space corroborates
this: LLM second-pass responses drift toward the model's
distributional prior, whereas human revisions remain localized
(Appendix~\ref{app:drift}).

\begin{figure*}[t]
\centering
\includegraphics[width=0.32\linewidth]{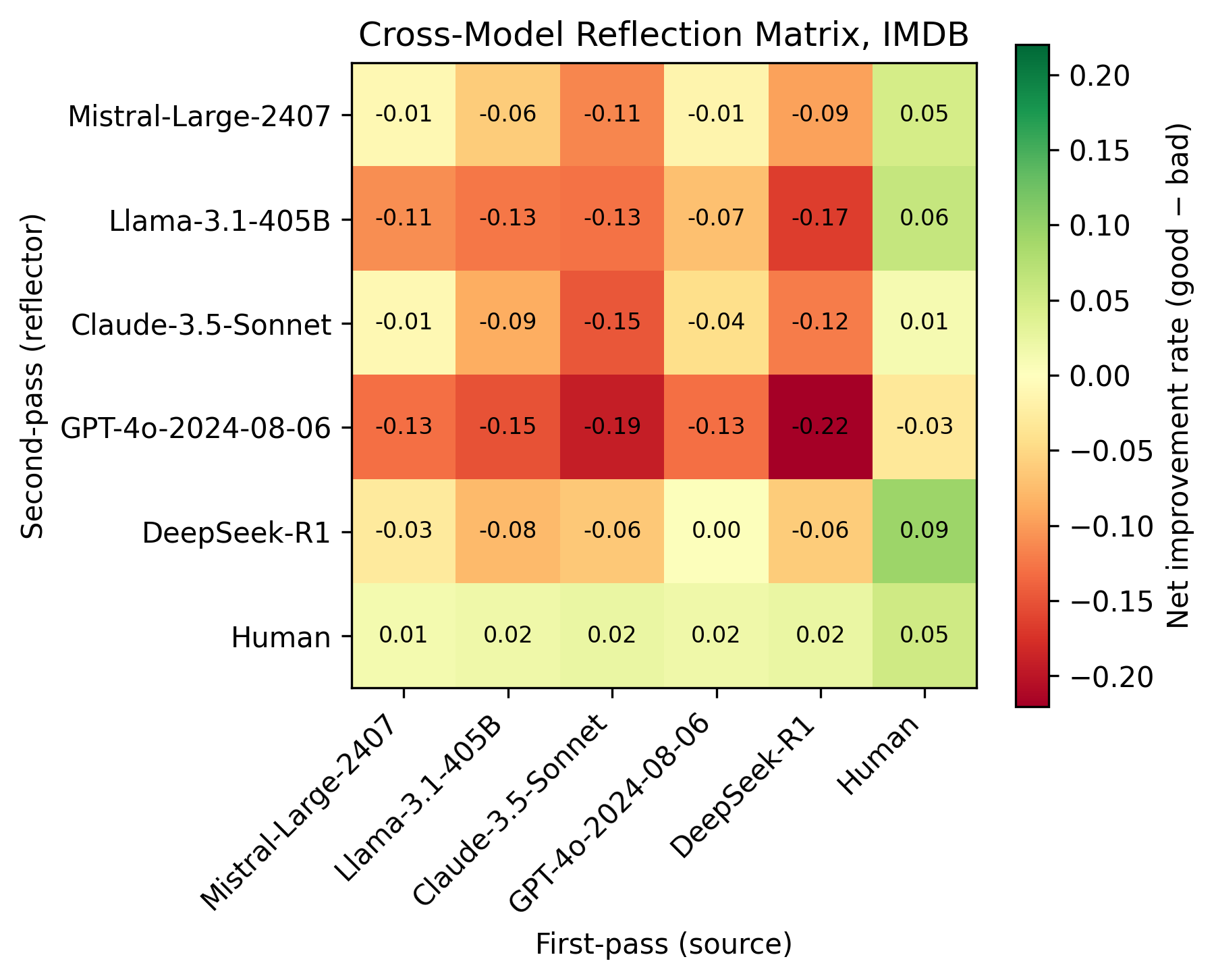}%
\hfill
\includegraphics[width=0.32\linewidth]{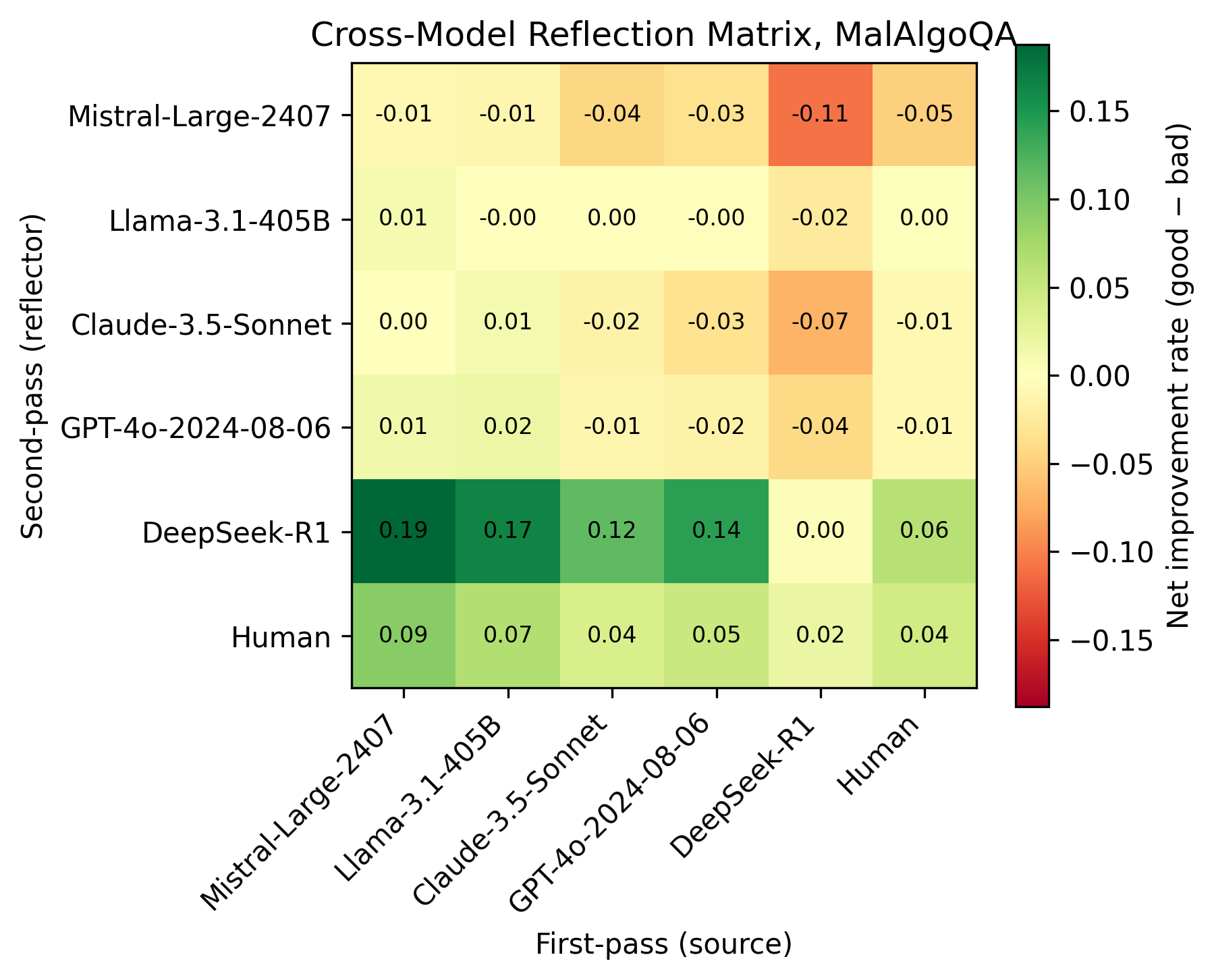}%
\hfill
\includegraphics[width=0.32\linewidth]{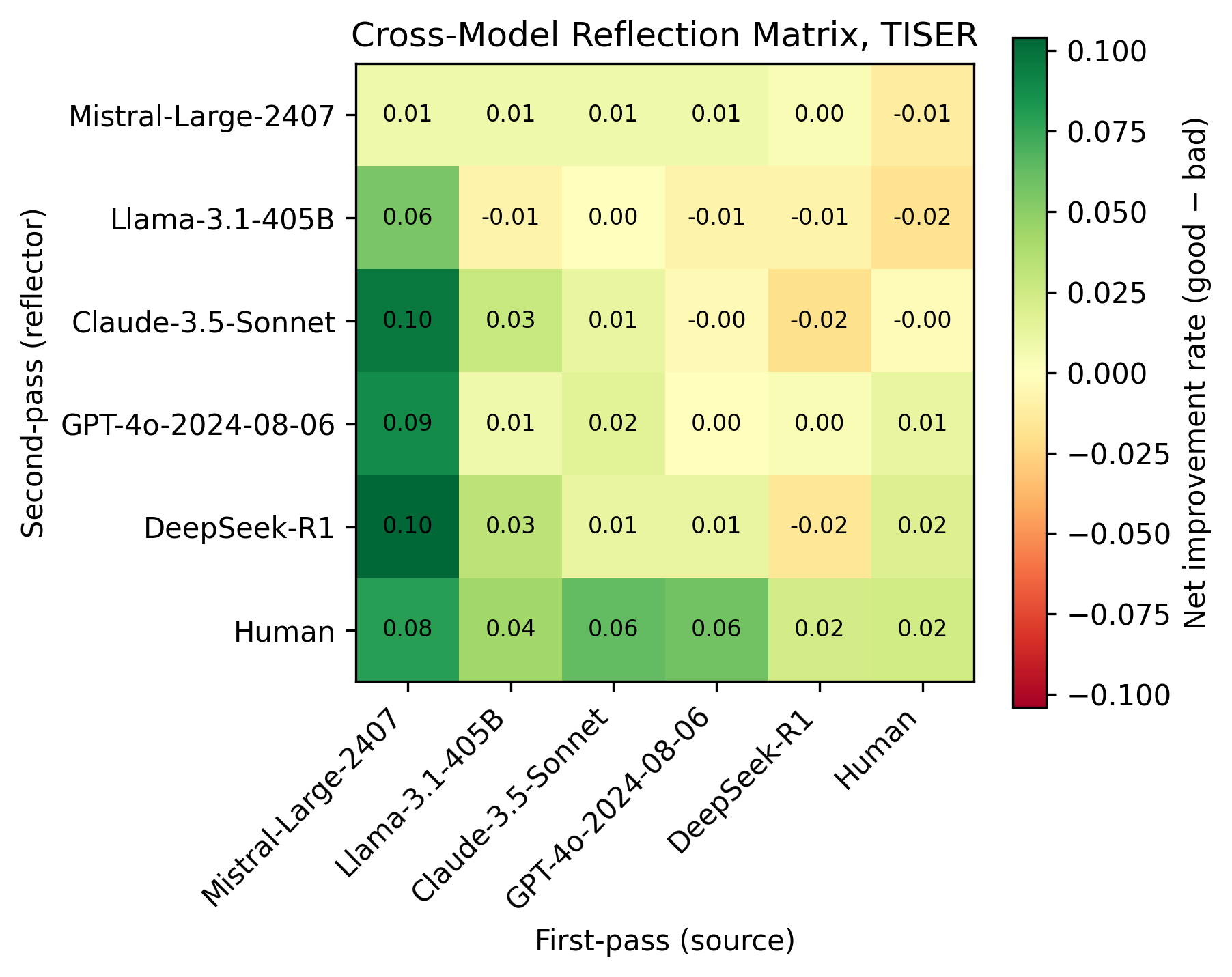}
\caption{Cross-model reflection matrices across the three evaluation
tasks: IMDb-Rating (left), MalAlgoQA (center), and TISER (right).
Rows denote reflectors (second pass), columns denote first-pass
sources. Each cell reports the net improvement rate (beneficial minus
harmful revision rate).}
\label{fig:cross-matrix}
\end{figure*}

\subsection{The Failure Lies in the Revision Step, Not the Input}
\label{sec:reviser-not-source}

The preceding analysis shows that LLM reflection fails to accumulate
information. Two explanations are possible: first-pass responses may
be too poor to build upon, or the revision mechanism may lack the
ability to identify which responses need correction. We examine
cross-agent revision outcomes from the HRF protocol
(Section~\ref{sec:hrf}) to distinguish between these
hypotheses.

Figure~\ref{fig:cross-matrix} reports the cross-model reflection
matrices for each task, where rows denote second-pass revisers and
columns denote first-pass sources. Each cell reports the net
improvement rate: the difference between beneficial and harmful
revisions.

Two patterns emerge. First, \textbf{human revisers produce
non-negative net improvement across nearly all source types and
tasks.} The human row is consistently non-negative regardless of
whether the first-pass response comes from a weak model, a strong
model, or another human. This holds even when the source output is
substantially stronger than the human's own first-pass performance
(e.g., LLM outputs on IMDb-Rating, where LLMs achieve lower MSE
than humans in the first pass).

Second, \textbf{LLM revision outcomes are source-dependent and
dominated by distributional attraction rather than error correction.}
On IMDb-Rating, the LLM--LLM cells are almost universally negative:
LLMs degrade each other's outputs despite those outputs being
reasonably strong. The LLM--Human cells are an apparent exception,
showing mostly positive values---but this reflects the fact that
human first-pass predictions on IMDb are weaker than LLM predictions
(Table~\ref{tab:main}), so pulling toward the model's prior happens
to reduce error. The improvement is incidental rather than diagnostic:
it tracks distributional proximity, not targeted correction. On
MalAlgoQA, the pattern is similar---only the strongest model
(DeepSeek-R1) acts as a broadly effective reviser, and its
effectiveness tracks the strength gap with the source rather than
error-specific reasoning.

Taken together, these results indicate that the failure lies in the
revision mechanism, not in source quality. Human revisers consistently
improve outputs regardless of source strength. LLM revisers, by
contrast, shift outputs toward their own distributional prior---a
process that helps when the source is weaker but harms when the
source is comparable or stronger. This locates the failure in the
revision step itself; it does not yet tell us which sub-step of
revision is responsible. We turn to that question next, measuring
error detection and error correction directly, and find that neither
is uniformly the bottleneck: the dominant sub-step varies by task and
by model.

\subsection{Error Detection Is Present but Task-Dependent}
\label{sec:detection}

To ask whether models detect their own errors, we measure the
per-run revision rate conditioned on first-pass correctness (K=10),
i.e.\ $P(\text{revise}\mid\text{wrong})$ versus
$P(\text{revise}\mid\text{correct})$. A model that detects its errors
should revise wrong answers more often than correct ones. On the
objective task MalAlgoQA this ratio is large (roughly $3$--$21\times$
across models), so a detection signal clearly exists; on the
subjective ordinal task IMDb it is weak (roughly $0.9$--$1.6\times$).
Detection is therefore task-dependent, not uniformly absent. Crucially,
where the signal exists it does not translate into improvement:
reflection gain on MalAlgoQA remains near zero across all five models
(Table~\ref{tab:main}) despite wrong answers being revised far more
often. The model flags candidates for revision but does not reliably
fix them---which motivates isolating the correction step directly.

\subsection{Error Correction Is Limited and Model-Dependent}
\label{sec:oracle}

To isolate correction from detection, we introduce
oracle-guided revision: a diagnostic experiment in which the model
receives a minimal external signal indicating whether its first-pass
answer was correct, without revealing the correct answer itself.

For each question, the oracle routes the model into one of two
conditions based on first-pass correctness:
\begin{itemize}
\item \textbf{Oracle-wrong}: ``Your previous answer is incorrect.
Please reconsider the question and provide a revised answer.''
\item \textbf{Oracle-correct}: ``Your previous answer is correct.
Please confirm or reconsider your answer.''
\end{itemize}

\vspace{-2mm}
This design isolates \emph{error detection}---knowing which answers
to revise---from \emph{error correction}---knowing what to revise
them to. The oracle provides exactly one bit of information per item
and never reveals the correct answer.

Table~\ref{tab:oracle} reports results on MalAlgoQA averaged over 10
runs. We report \emph{recovery rate} (fraction of incorrect
first-pass answers corrected) and \emph{retention rate} (fraction of
correct first-pass answers preserved). Because MalAlgoQA is a
4-choice task, a model that excludes its original answer and guesses
uniformly achieves a recovery rate of $33.3\%$.

\begin{table}[t]
\centering
\caption{Oracle-guided revision on MalAlgoQA. Recovery rate: fraction
of incorrect first-pass answers corrected. Retention rate: fraction of
correct first-pass answers preserved. Random re-guess baseline:
$0.333$. All values averaged over 10 runs.}
\label{tab:oracle}
\small
\setlength{\tabcolsep}{2.5pt}
\begin{tabular}{lccccc}
\toprule
Condition & Mistral-Large & Llama-3.1-405B & Claude-3.5 & GPT-4o & DeepSeek-R1 \\
\midrule
1st pass
  & 0.750 & 0.776 & 0.829 & 0.801 & 0.948 \\
Self-reflection
  & 0.741\tiny{$\pm$.005} & 0.757\tiny{$\pm$.012}
  & 0.807\tiny{$\pm$.003} & 0.791\tiny{$\pm$.004}
  & 0.954\tiny{$\pm$.004} \\
Oracle-guided
  & 0.776\tiny{$\pm$.003} & 0.816\tiny{$\pm$.004}
  & 0.872\tiny{$\pm$.003} & 0.865\tiny{$\pm$.004}
  & 0.965\tiny{$\pm$.003} \\
\midrule
Recovery rate
  & 0.171\tiny{$\pm$.010} & 0.168\tiny{$\pm$.017}
  & 0.372\tiny{$\pm$.016} & 0.334\tiny{$\pm$.022}
  & 0.413\tiny{$\pm$.057} \\
Retention rate
  & 0.978\tiny{$\pm$.003} & 0.999\tiny{$\pm$.001}
  & 0.973\tiny{$\pm$.002} & 0.997\tiny{$\pm$.001}
  & 0.995\tiny{$\pm$.002} \\
\bottomrule
\end{tabular}
\end{table}

Oracle-guided revision improves accuracy over self-reflection for all
five models, with gains ranging from $+1.1$ points (DeepSeek-R1) to
$+7.4$ (GPT-4o). But the recovery rate---how often a flagged-wrong
answer is actually corrected---is \emph{model-dependent} and does not
support a uniform ``correction is intact'' reading. Recovery exceeds
the $33.3\%$ random-reshuffle baseline only for the stronger models
(DeepSeek-R1: $41.3\%$, Claude: $37.2\%$), sits at chance for GPT-4o
($33.4\%$), and falls well below it for the weaker models (Mistral:
$17.1\%$, Llama: $16.8\%$). If correction capacity were intact and
only detection missing, a flagged-wrong answer should recover at least
at the reshuffle baseline; for most models it does not. What is
consistent across models is \emph{retention}: the oracle reliably
prevents correct$\rightarrow$wrong changes ($97$--$99.9\%$), which is
where much of its accuracy gain comes from, especially for weaker
models.

Taken with the detection measure above, the picture is one of
heterogeneity rather than a single bottleneck. Detection is present on
objective tasks but weak on subjective ones; correction, as unlocked
by an oracle signal, is genuine for the strongest models but limited
for weaker ones, whose main benefit from the oracle is retention. We
therefore do not localize the failure to a single sub-step. The
consistent finding is structural: a binary external signal is enough
to change behavior at all (unlike self-conditioning), but often
insufficient to overwrite a confident wrong prior. While a perfect
oracle is unavailable in practice, approximate error signals are
readily obtainable through tool-based verification, verifier models, or
lightweight human feedback~\citep{cobbe2021training, lightman2023let},
suggesting that such mechanisms are a more promising investment than
increasing reflection depth or prompt complexity.

Multi-iteration experiments (up to 5 passes) using iterative
self-revision, self-ensemble, and cross-model strategies further
confirm that incremental information gain collapses to near zero
after the first revision step across all configurations
(Appendix~\ref{app:diminishing}).

\section{Related Work}
\label{sec:related}

\subsection{Metacognition and Human Revision}

The study of how humans monitor and revise their reasoning has a long
history in cognitive psychology.
\citet{flavell1979metacognition} introduced the concept of
metacognition, and \citet{nelson1990metamemory} formalized it as a
two-level system in which a meta-level process monitors and controls
object-level cognition. Subsequent work has shown that effective
revision depends on monitoring accuracy---the ability to distinguish
correct from incorrect responses---rather than on revision
volume~\citep{dunlosky2008metacognition, veenman2006metacognition}.
Error monitoring studies further demonstrate that humans detect and
signal their own mistakes during task performance, driving adaptive
learning~\citep{yeung2012metacognition}.

We use human annotators as a behavioral reference for what effective
revision looks like under our protocol, not as a claim about human
metacognition: our non-expert crowdworker sample is not designed to
measure monitoring capacity in the sense of~\citet{nelson1990metamemory}.
At the behavioral level, human annotators in our study produce sparse,
targeted revisions (Section~\ref{sec:revision-precision}) with positive
information gain across both objective and subjective tasks
(Sections~\ref{sec:two-failures-main}), whereas LLM revision does not
improve under identical conditions. For LLMs, our detection and oracle
experiments (Sections~\ref{sec:detection},~\ref{sec:oracle}) show that
which sub-step of revision is deficient varies by task and by model.

\subsection{Reflection and Self-Correction in LLMs}
Reflexion~\citep{shinn2023reflexion} uses verbal self-reflection as a
reinforcement signal for sequential decision-making; subsequent work
applies similar ideas to hallucination
mitigation~\citep{ji2023towards} and general problem
solving~\citep{renze2024self}. However, the reliability of intrinsic
self-correction remains contested: \citet{huang2023large} show that
LLMs cannot self-correct reasoning without external feedback, confirmed
by \citet{kamoi2024can} and \citet{li2024selective}. Our work adds a matched
human baseline, characterizes two distinct failure modes (neutral
re-generation on objective tasks, distributional flattening on
subjective tasks), and shows via direct detection and oracle
experiments (Sections~\ref{sec:detection},~\ref{sec:oracle}) that the
dominant sub-step of the failure varies by task and by model rather
than reducing to a single bottleneck. Our use of ``revision'' follows
this prompting literature rather than the formal belief-revision
tradition (AGM;~\citealp{alchourron1985logic, darwiche1997logic}),
where revision requires triggering information that conflicts with the
agent's prior state; self-conditioned re-prompting supplies no such
information, which is consistent with our finding that it behaves as
re-generation.

\subsection{Human--LLM Comparative Studies}

\citet{jones2024comparing} evaluate LLMs on theory-of-mind tasks
calibrated against human performance. \citet{yax2024studying} compare
human and LLM reasoning on cognitive tasks. \citet{kamoi2024can}
contrast human and LLM error patterns in self-correction settings.
These studies evaluate single-pass responses rather than revision
dynamics. To our knowledge, no prior work has (i)~compared human and
LLM revision under a unified multi-pass protocol, (ii)~included
cross-agent revision, or (iii)~provided an information-theoretic
characterization of revision dynamics. HRF addresses all three gaps.

\section{Discussion and Conclusion}
\label{sec:discussion}

\paragraph{Reflection as conditioned re-generation.}
Our information-theoretic analysis reveals that LLM ``reflection''
is better understood as conditioned re-generation: without an
external error signal, self-conditioned revision cannot reduce
uncertainty about the target, yielding neutral re-sampling on
objective tasks ($\Delta I \approx 0$) and distributional flattening
on subjective tasks ($\Delta I < 0$).

\paragraph{Where revision fails varies by task and model.}
The monitoring-and-control framework from cognitive
science~\citep{nelson1990metamemory} distinguishes error detection from
error correction. Our direct measurements show that neither is
uniformly the bottleneck. Self-error detection is present on objective
multiple-choice tasks (wrong answers are revised several times more
often than correct ones) but weak on subjective ordinal judgments.
Correction, as unlocked by an oracle error signal, exceeds a
random-reshuffle baseline for the strongest models but falls below it
for weaker ones, whose oracle benefit comes mainly from retention.
Rather than a single dissociation, the failure is heterogeneous across
tasks and models, unified only by the structural point below.

\paragraph{Implications for system design.}
Our findings suggest three design principles for effective revision
systems:
\begin{enumerate}
\item \emph{External verification is necessary.} Self-conditioned
revision without new information cannot accumulate task-relevant signal.
Effective reflection requires an independent source of error
detection---whether from a separate verifier model, tool-based
checking, or human feedback.

\item \emph{Weak--strong asymmetry can be exploited.} Improvements are
more likely when a stronger agent revises a weaker response
(Section~\ref{sec:reviser-not-source}), suggesting that multi-agent
architectures with heterogeneous capabilities may be more effective than
single-model self-revision.

\item \emph{Iteration depth should be bounded.} Diminishing returns
across iterations (Appendix~\ref{app:diminishing}) imply that
computational resources are better allocated to independent sampling or
external verification than to repeated self-conditioning.
\end{enumerate}

\paragraph{Limitations.}
Our information-theoretic analysis is restricted to tasks with finite
answer spaces and reliable ground truth (MalAlgoQA and IMDb-Rating).
Extending these measures to open-ended generation remains an open
challenge. Human annotators are non-expert crowdworkers; expert
annotators might exhibit different revision patterns, though the
structural differences we observe (sparse, directional revision versus
high-variance re-sampling) are unlikely to depend on domain expertise
alone. Our second-pass prompt asks the model to agree or disagree with
the prior answer, which may itself encourage revision. As a control, we
re-ran the IMDb second pass with a neutral prompt that asks the model to
revise only upon identifying a specific error; this sharply reduces the
revision rate and attenuates, though does not eliminate, the negative
reflection gain, indicating that part of the IMDb effect is
prompt-induced disagreement while part reflects genuine drift away from
the target. This also means our revision-rate magnitudes are
prompt-dependent and should be read as behavior under an
agree-or-disagree instruction rather than as intrinsic constants.
Finally, our oracle experiments use a binary error signal; more nuanced
forms of feedback (e.g., identifying the type of error) may yield
additional insights.

\paragraph{Conclusion.}
We introduced the Human--LLM Reflection Framework (HRF) and identified
two failure modes of LLM reflection through per-iteration
information-theoretic analysis: neutral re-generation on objective tasks
($\Delta I \approx 0$) and distributional flattening on subjective tasks
($\Delta I < 0$). Human revision, by contrast, yields positive
information gain across both settings. Diagnostic experiments localize
the failure to the revision step rather than input quality, and show
that which sub-step dominates---detecting which answers to revise
versus producing a better one---varies by task and by model rather than
reducing to a single mechanism. The unifying point is structural:
without external information, self-conditioned revision cannot reduce
uncertainty about the target. These results reframe LLM reflection as
conditioned re-generation and point toward revision architectures
grounded in external verification rather than self-conditioned
prompting.

\section*{LLM Usage Disclosure} 
Large language models are central to this work in two capacities. First, as experimental subjects: five models (Llama-3.1-405B, Claude-3.5-Sonnet, Mistral-Large-0724, GPT-4o-20240806, and DeepSeek-R1, with GPT-5-mini and GPT-5.2 in supplementary experiments) were evaluated under the HRF protocol to study reflection behavior. Second, as an evaluation tool: a fixed LLM judge~\citep{alexandru2025atla} was used to assess correctness of free-form responses on the TISER benchmark, as described in Section~\ref{sec:hrf}.

\bibliography{colm2026_conference}
\bibliographystyle{colm2026_conference}

\appendix
\section{Annotation Details}
\label{app:annotation-details}

All human annotators are English-speaking, based in India, with at
least high school--level education. Across tasks, we employ
approximately 10--15 unique annotators per dataset. The design targets
5 annotators per item, with the final average ranging between 4 and 5
due to occasional missed or invalid submissions. Quality control
includes removal of incomplete submissions and validation of required
output formats.

\begin{figure*}[t]
\centering
\includegraphics[width=\textwidth]{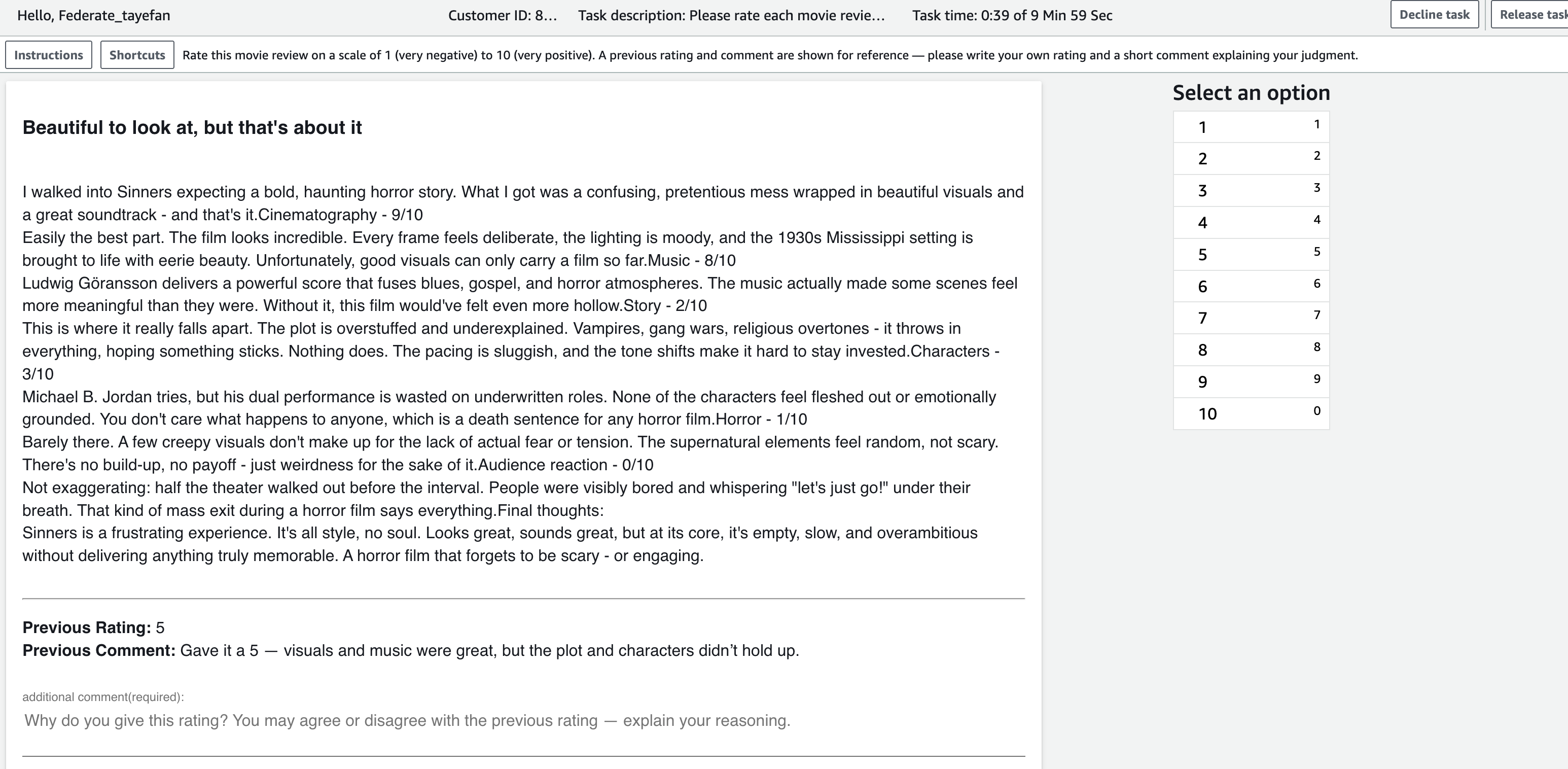}
\caption{Example annotation UI in Amazon SageMaker Ground Truth for
the IMDb-Rating task.}
\label{fig:ui-imdb}
\end{figure*}

\begin{figure*}[t]
\centering
\includegraphics[width=\textwidth]{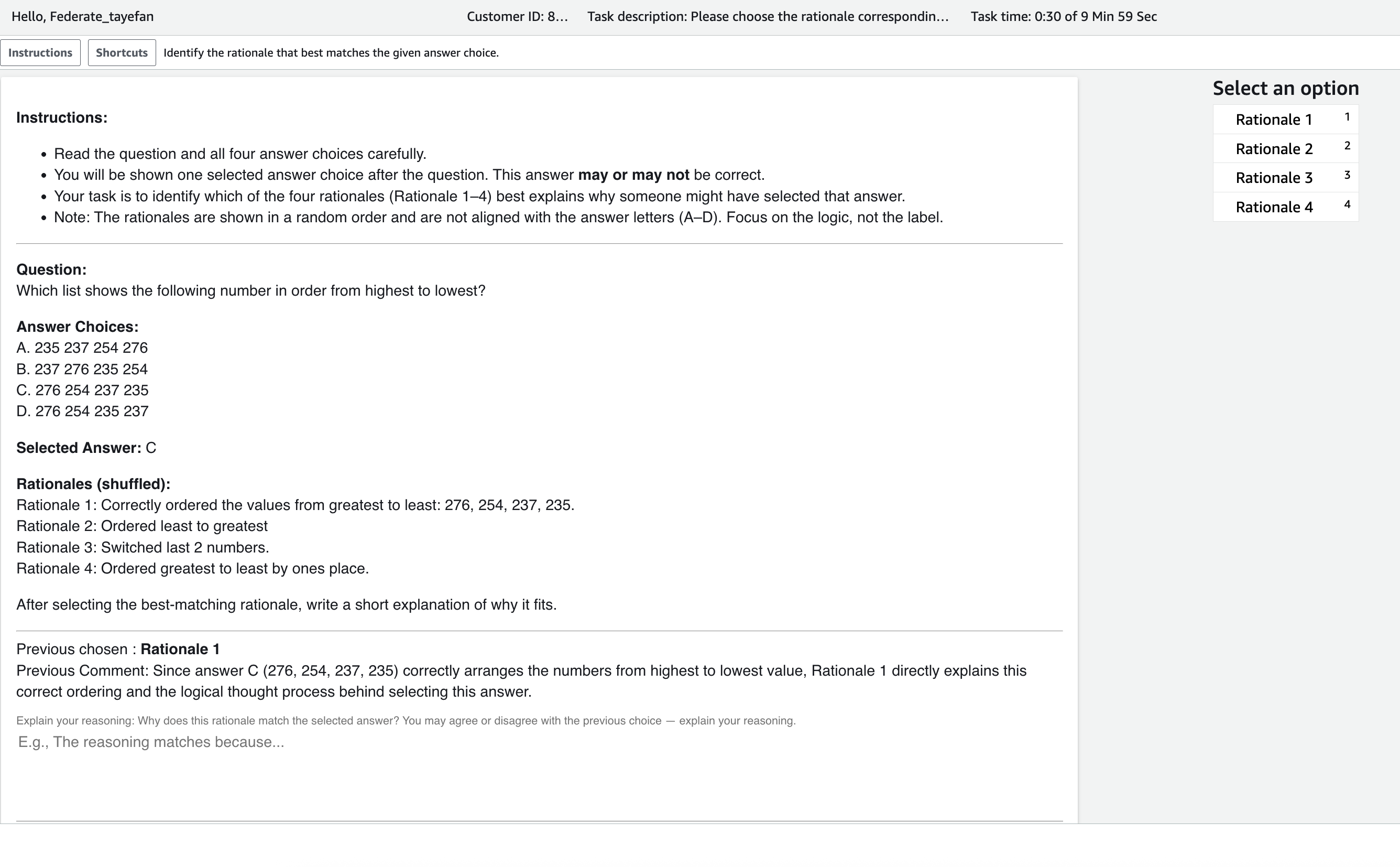}
\caption{Example annotation UI in Amazon SageMaker Ground Truth for
the MalAlgoQA task.}
\label{fig:ui-malalgoqa}
\end{figure*}

\begin{figure*}[t]
\centering
\includegraphics[width=\textwidth]{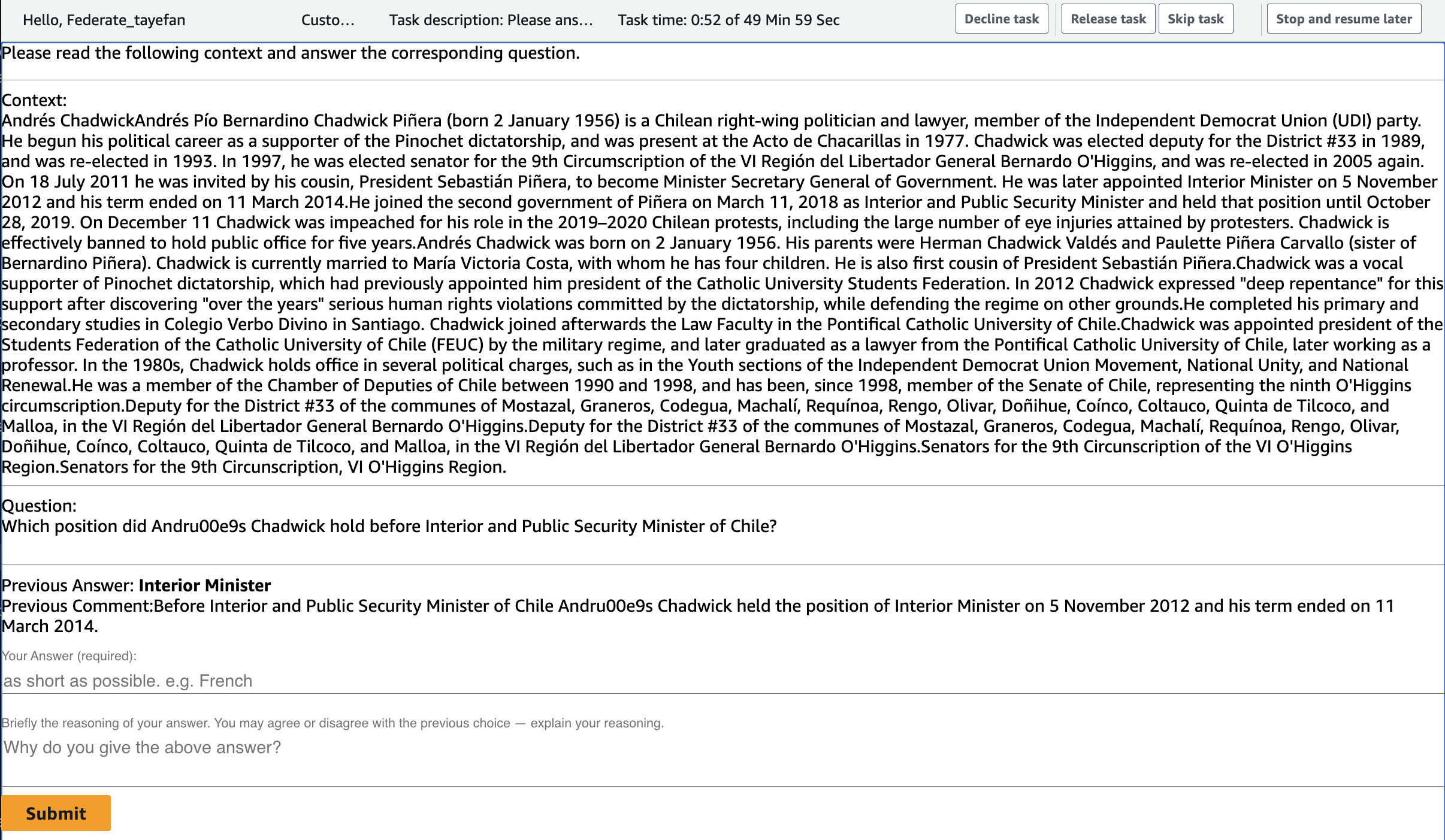}
\caption{Example annotation UI in Amazon SageMaker Ground Truth for
the TISER task.}
\label{fig:ui-tiser}
\end{figure*}

We built custom annotation UIs using liquid templates in Amazon
SageMaker Ground Truth. Example interfaces are shown in
Figures~\ref{fig:ui-imdb}--\ref{fig:ui-tiser}.

\paragraph{Inter-annotator agreement.}
Table~\ref{tab:iaa} reports Krippendorff's $\alpha$ for each task
and pass. For IMDb-Rating, we use the ordinal variant of $\alpha$ to
account for the ordered 1--10 scale; for MalAlgoQA and TISER, we use
the nominal variant. TISER responses are free-form; we map them to a
ternary label set (\texttt{correct}/\texttt{incorrect}/\texttt{unknown})
before computing agreement. The second-pass column includes both
self-revision and peer-revision annotations, roughly doubling the
average number of annotations per item.

Agreement is moderate to high across all settings. IMDb-Rating shows
the strongest agreement ($\alpha = 0.819$ first pass, $0.892$ second
pass), consistent with the ordinal structure of the task. MalAlgoQA
exhibits substantial agreement ($\alpha = 0.685 \to 0.731$) with
near-ceiling majority agreement ($92.4\% \to 96.6\%$), indicating
that most items admit a clear consensus answer. TISER yields the
lowest $\alpha$ values ($0.529 \to 0.560$), reflecting the inherent
ambiguity of free-form temporal reasoning even after ternary mapping;
this is consistent with human first-pass accuracy on TISER being the
lowest across all three tasks (Table~\ref{tab:main}). Across all
three tasks, both $\alpha$ and majority agreement increase from the
first to the second pass, suggesting that revision produces mild
convergence among annotators---consistent with the positive reflection
gains reported in Section~\ref{sec:empirical}.

\begin{table}[h]
\centering
\small
\caption{Inter-annotator agreement (Krippendorff's $\alpha$) by task
and pass. IMDb-Rating uses ordinal $\alpha$; MalAlgoQA and TISER use
nominal $\alpha$ (TISER responses mapped to
correct/incorrect/unknown). Majority\% is the fraction of items on
which a strict majority of annotators agree.}
\label{tab:iaa}
\begin{tabular}{llcccc}
\toprule
Task & Pass & Metric & $\alpha$
     & Avg ann/item & Majority\% \\
\midrule
IMDb-Rating & 1st & ordinal & 0.819 & 4.47 & 57.4\% \\
IMDb-Rating & 2nd & ordinal & 0.892 & 8.49 & 66.7\% \\
\midrule
MalAlgoQA   & 1st & nominal & 0.685 & 4.37 & 92.4\% \\
MalAlgoQA   & 2nd & nominal & 0.731 & 9.33 & 96.6\% \\
\midrule
TISER       & 1st & nominal & 0.529 & 3.33 & 90.4\% \\
TISER       & 2nd & nominal & 0.560 & 7.08 & 92.4\% \\
\bottomrule
\end{tabular}
\end{table}

\clearpage

\newtcolorbox{promptbox}[1][]{
  colback=gray!5, colframe=gray!50,
  fonttitle=\bfseries\small, title={#1},
  boxrule=0.5pt, arc=2pt, left=4pt, right=4pt,
  top=4pt, bottom=4pt
}

\definecolor{pass2}{RGB}{0,80,160}
\section{Prompt Templates}
\label{app:prompts}

Each task uses a two-pass prompt structure. The first-pass prompt contains the base task instructions. The second-pass prompt extends it with a reflection preamble (shown in {\color{pass2}\textbf{blue}}) that presents the prior annotator's response and asks the model to agree or disagree. All prompts are used identically for human and LLM annotators.

\begin{promptbox}[IMDb-Rating]
\ttfamily\footnotesize\raggedright
You are an experienced movie reviewer.
{\color{pass2}You will be given an IMDB movie review, along with
another annotator's review rating and explanation. Carefully read the
review, then consider the annotator's rating and explanation. Based on
your own judgment, either agree or disagree with the annotator's
rating.}
Read the provided IMDB movie review carefully and assign a sentiment
rating from 1 to 10, where 1 means extremely negative and 10 means
extremely positive. Write a short comment (1--2 sentences) explaining
your reasoning.\par
{\color{pass2}Do not simply copy the annotator's explanation. Agreement
is not required.}\par
Respond in this JSON format:\par
\{"rating": <integer>, "comment": "<your explanation>"\}
\end{promptbox}

\begin{promptbox}[MalAlgoQA]
\ttfamily\footnotesize\raggedright
You are an assistant for a math QA evaluation task. You will be given a
math question, four answer choices (A--D), four rationales (R1--R4),
and a selected answer (e.g., `B'){\color{pass2}, and another
annotator's chosen rationale with explanation}.\par
Your task is to {\color{pass2}read the question and selected answer
yourself, then consider the annotator's rationale and explanation.
Based on your own judgment, either agree or disagree.} Identify which
rationale (R1, R2, R3, or R4) best matches the given selected answer,
and provide a short comment (1--2 sentences) explaining your
reasoning.\par
Important:\par
-- The selected answer may be incorrect. Your task is to find the
rationale that most likely supports the *selected* answer.\par
-- Rationales R1--R4 are not necessarily aligned with answer options
A--D by position. You must determine the correct match based on
content.\par
{\color{pass2}-- Do not simply copy the previous explanation. Agreement
is not required.}\par
Respond in this JSON format:\par
\{"rationale": "<R1|R2|R3|R4>", "comment": "<brief explanation>"\}
\end{promptbox}

\begin{promptbox}[TISER]
\ttfamily\footnotesize\raggedright
You are an assistant for a temporal reasoning evaluation task. You will
be given a context describing events and a question about the temporal
order or relationship of those events{\color{pass2}, and another
annotator's answer with reasoning}.\par
Your task:\par
1. Carefully read the context and understand the sequence of
events.\par
{\color{pass2}2. Review the annotator's answer and reasoning.\par
3. Based on your own judgment, either agree or disagree. Provide your
own answer and reasoning. Do not simply copy the previous
explanation.\par}%
Important:\par
-- Your answer must be based only on the provided context.\par
{\color{pass2}-- Agreement with the previous annotator is not
required.}\par
-- Provide a short explanation (1--2 sentences) that makes your
reasoning clear.\par
-- If the question is not answerable with the context, put "unknown"
in your answer.\par
Respond in this JSON format:\par
\{"answer": "<your answer>", "comment": "<your explanation>"\}
\end{promptbox}

\section{Self- vs.\ Peer-Revision Plots}
\label{app:self-peer-plots}

This appendix reports revision outcomes for all models (including humans) under two revision types: \emph{self-revision} (revising one's own first-pass answer) and \emph{peer-revision} (revising an answer produced by another annotator). In each subplot, the x-axis enumerates models, and the y-axis reports the fraction of revision outcomes falling into four categories: self-revision leading to a beneficial change, peer-revision leading to a beneficial change, self-revision leading to a harmful change, and peer-revision leading to a harmful change.

\begin{figure*}[t]
\centering
\includegraphics[width=0.32\linewidth]{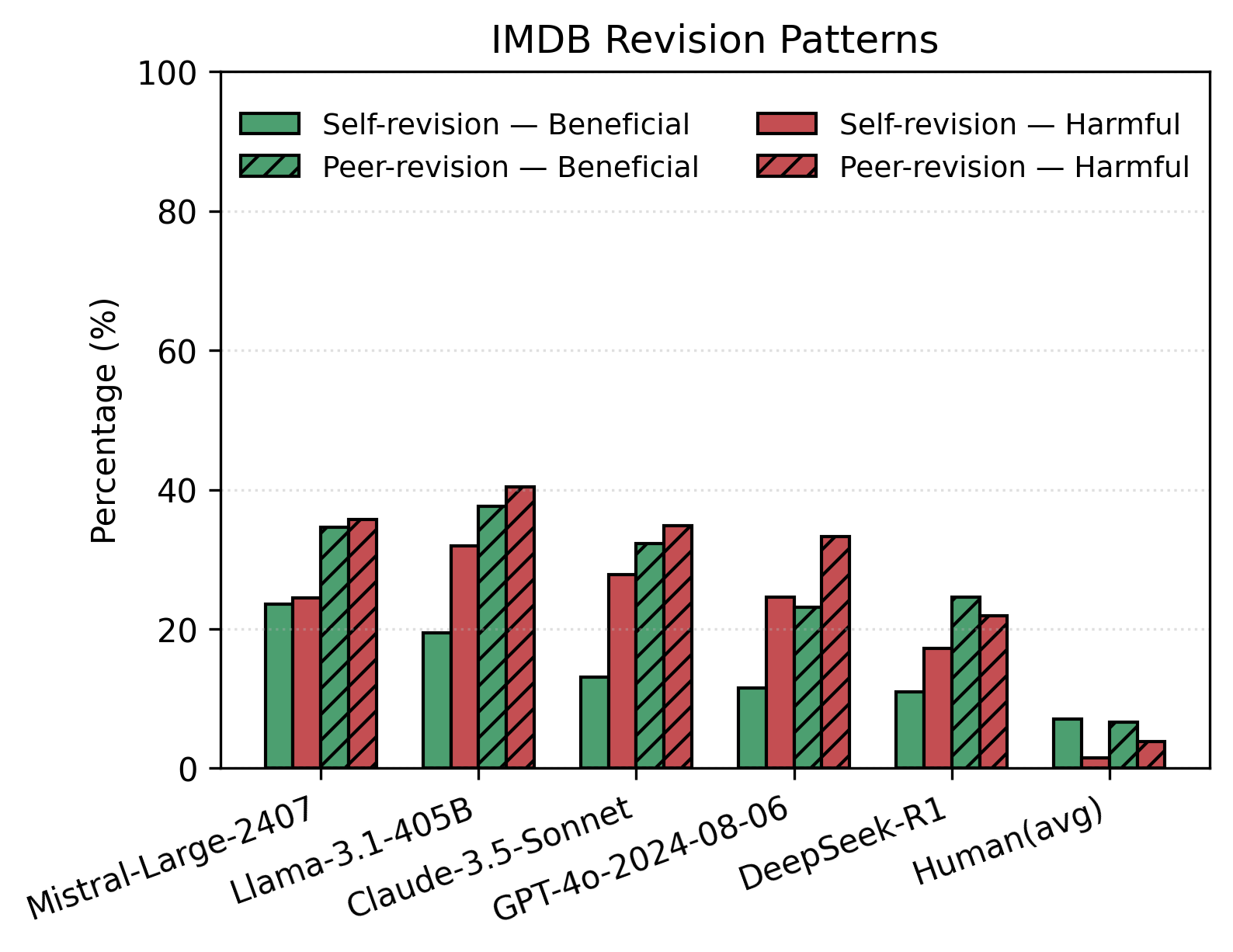}%
\hfill
\includegraphics[width=0.32\linewidth]{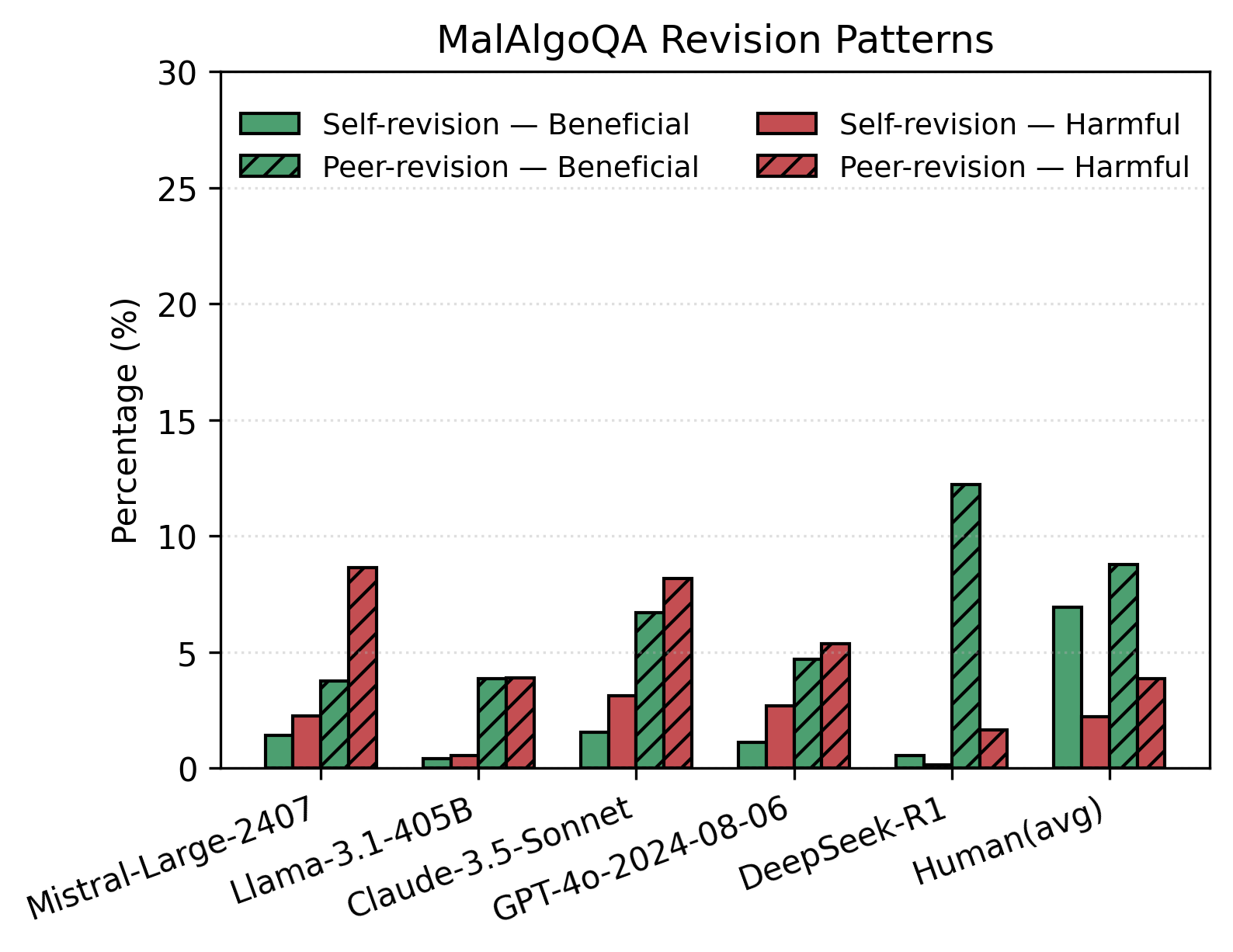}%
\hfill
\includegraphics[width=0.32\linewidth]{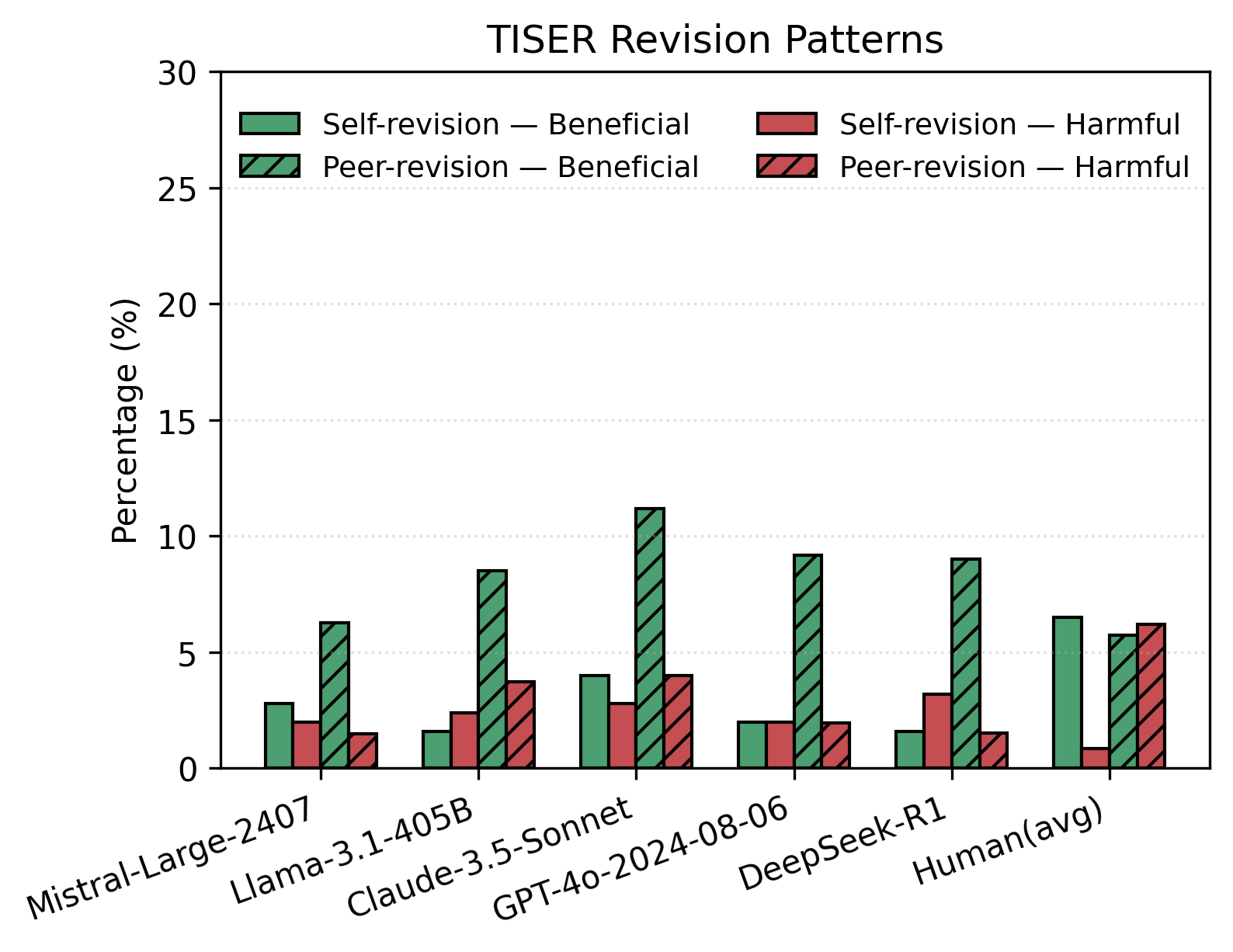}
\caption{Self- and peer-revision outcomes on IMDb-Rating (left),
MalAlgoQA (center), and TISER (right).}
\label{fig:self-peer-all}
\end{figure*}

\section{Statistical Testing Details}
\label{app:stat-tests}

To assess whether reflection provides systematic improvements beyond
stochastic decoding variability, we perform repeated-run statistical
testing using paired comparisons. For each model--task pair, the full
evaluation is repeated $K=10$ times under two second-pass conditions:
(i)~\emph{re-generation}, where the model answers the dataset again
without access to its first-pass output, and (ii)~\emph{reflection},
where the model conditions on its own first-pass output following the
reflection protocol (Section~\ref{sec:hrf}). Each repetition yields a
run-level performance value (e.g., MSE or accuracy), producing two
paired sets of $K$ measurements. We compute paired differences
\[
d_k = \mathrm{Metric}^{(k)}_{\mathrm{refl}}
    - \mathrm{Metric}^{(k)}_{\mathrm{re\text{-}gen}},
\]
and apply a standard two-sided paired
$t$-test~\citep{dietterich1998approximate, bouckaert2004evaluating} to
evaluate whether conditioning on the first-pass output produces a
systematic performance change relative to re-generation. For each
model--task pair, we report the mean paired change~$\Delta$, the
associated $t$-statistic, two-sided $p$-value, and a 95\% confidence
interval, along with the rejection decision at $\alpha = 0.05$.

Table~\ref{tab:paired-ttest} reports the complete results. Across all
settings, we do not observe any case in which reflection yields a
statistically significant \emph{positive} improvement over stochastic
re-generation.

\begin{table*}[t]
\centering
\small
\setlength{\tabcolsep}{3pt}
\begin{tabular}{llccccccc}
\toprule
& Model
& $\mathrm{Metric}_{\mathrm{re\text{-}gen}}$
& $\mathrm{Metric}_{\mathrm{refl}}$
& $\Delta$ (\%)
& $t$
& $p$-value
& $CI_{95\%}$
& $H_0$ rej.? \\
\midrule
\multirow{6}{*}{\rotatebox[origin=c]{90}{IMDb}}
& \multicolumn{8}{l}{\textit{Metric: MSE (lower is better
  $\downarrow$)}} \\
& Mistral-Large   & 1.747 & 1.884
  & \textcolor{red}{$-$7.9}   & 17.86 & $2.46\!\times\!10^{-8}$
  & (0.120, 0.155)        & \cmark \\
& Llama-3.1-405B  & 1.967 & 2.240
  & \textcolor{red}{$-$13.9}  & 18.05 & $2.24\!\times\!10^{-8}$
  & (0.238, 0.307)        & \cmark \\
& Claude-3.5      & 1.592 & 2.048
  & \textcolor{red}{$-$28.7}  & 54.03 & $1.28\!\times\!10^{-12}$
  & (0.437, 0.475)        & \cmark \\
& GPT-4o          & 1.742 & 2.013
  & \textcolor{red}{$-$15.5}  & 19.11 & $1.36\!\times\!10^{-8}$
  & (0.239, 0.303)        & \cmark \\
& DeepSeek-R1     & 1.814 & 1.958
  & \textcolor{red}{$-$7.9}   & 8.08  & $2.05\!\times\!10^{-5}$
  & (0.104, 0.184)        & \cmark \\
\midrule
\multirow{6}{*}{\rotatebox[origin=c]{90}{MalAlgoQA}}
& \multicolumn{8}{l}{\textit{Metric: Accuracy (higher is better
  $\uparrow$)}} \\
& Mistral-Large   & 0.738 & 0.742
  & \textcolor{teal}{+0.6}    & 2.03  & $7.26\!\times\!10^{-2}$
  & ($-$0.001, 0.009)     & \xmark \\
& Llama-3.1-405B  & 0.767 & 0.762
  & \textcolor{red}{$-$0.6}   & $-$0.72 & $4.87\!\times\!10^{-1}$
  & ($-$0.020, 0.010)     & \xmark \\
& Claude-3.5      & 0.829 & 0.809
  & \textcolor{red}{$-$2.4}   & $-$11.83 & $8.67\!\times\!10^{-7}$
  & ($-$0.024, $-$0.016)  & \cmark \\
& GPT-4o          & 0.794 & 0.792
  & \textcolor{red}{$-$0.2}   & $-$0.43 & $6.76\!\times\!10^{-1}$
  & ($-$0.009, 0.006)     & \xmark \\
& DeepSeek-R1     & 0.957 & 0.954
  & \textcolor{red}{$-$0.4}   & $-$1.75 & $1.15\!\times\!10^{-1}$
  & ($-$0.008, 0.001)     & \xmark \\
\midrule
\multirow{6}{*}{\rotatebox[origin=c]{90}{TISER}}
& \multicolumn{8}{l}{\textit{Metric: Accuracy (higher is better
  $\uparrow$)}} \\
& Mistral-Large   & 0.471 & 0.475
  & \textcolor{teal}{+0.9}    & 2.18  & $5.71\!\times\!10^{-2}$
  & ($-$0.000, 0.009)     & \xmark \\
& Llama-3.1-405B  & 0.494 & 0.490
  & \textcolor{red}{$-$0.8}   & $-$1.37 & $2.04\!\times\!10^{-1}$
  & ($-$0.011, 0.003)     & \xmark \\
& Claude-3.5      & 0.530 & 0.548
  & \textcolor{teal}{+3.5}    & 4.36  & $1.84\!\times\!10^{-3}$
  & (0.009, 0.028)        & \cmark \\
& GPT-4o          & 0.526 & 0.528
  & \textcolor{teal}{+0.3}    & 0.49  & $6.37\!\times\!10^{-1}$
  & ($-$0.006, 0.009)     & \xmark \\
& DeepSeek-R1     & 0.540 & 0.536
  & \textcolor{red}{$-$0.7}   & $-$1.87 & $9.51\!\times\!10^{-2}$
  & ($-$0.009, 0.001)     & \xmark \\
\bottomrule
\end{tabular}
\caption{Paired run-level statistical tests over $K=10$ repeated runs
per model and task. $\Delta$ is the paired relative change in
percentage points. $CI_{95\%}$ reports the confidence interval on the
raw metric difference. $H_0$ rejected indicates $p < 0.05$.}
\label{tab:paired-ttest}
\end{table*}

\section{Geometric Evidence for Distributional Attraction}
\label{app:drift}

To assess whether reflection produces targeted error correction or
instead yields a conditioned variant of a model's typical output, we
analyze reflection as geometric drift in embedding space relative to
an item-conditioned task prior.

For each question $q$, we obtain a set of $K$ unconditioned
second-pass reasoning samples generated without access to the
first-pass response. We take their mean embedding
\[
\mu_q = \frac{1}{K}\sum_{k=1}^{K}
        \mathbf{e}^{(k)}_{q,\mathrm{regen}}
\]
as a local approximation of the model's intrinsic reasoning
distribution for that item. Given the first-pass reasoning embedding
$\mathbf{e}^{(1)}_q$ and the second-pass (reflection) embedding
$\mathbf{e}^{(2)}_q$, we compute their cosine distances to the
estimated prior,
\[
d^{(1)}_q = \mathrm{dist}(\mathbf{e}^{(1)}_q,\, \mu_q), \qquad
d^{(2)}_q = \mathrm{dist}(\mathbf{e}^{(2)}_q,\, \mu_q),
\]
and define the per-question drift as
$\Delta_q = d^{(2)}_q - d^{(1)}_q$. Negative values indicate that
reflection moves the output closer to the model's task prior;
positive values indicate movement away from it.

\begin{figure*}[t]
\centering
\includegraphics[width=0.88\linewidth]{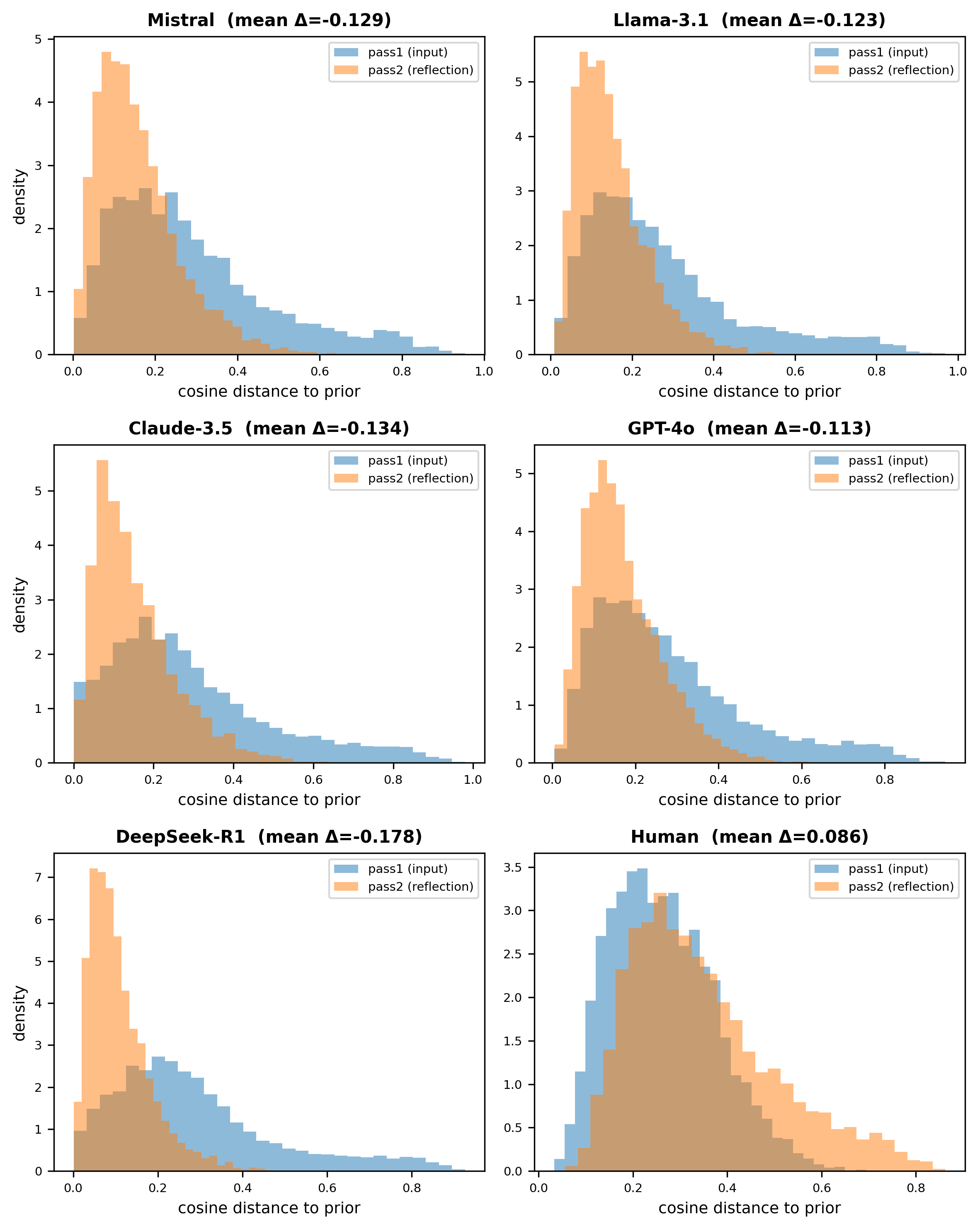}
\caption{Distributions of cosine distance to the estimated task
prior for first-pass (blue) and reflection (orange) responses on
MalAlgoQA. Each panel reports the mean per-question drift
$\Delta = d^{(2)}_q - d^{(1)}_q$. All five LLMs
show negative drift (reflection moves toward the prior); human
annotators show slight positive drift, indicating no systematic
prior-aligned movement.}
\label{fig:drift-main}
\end{figure*}

Figure~\ref{fig:drift-main} visualizes the distributions of
$d^{(1)}_q$ (first pass) and $d^{(2)}_q$ (reflection) for all five
LLMs and human annotators on MalAlgoQA. Each panel reports the mean
drift $\bar{\Delta}$ in its title.

All five LLMs exhibit a consistent pattern: the reflection
distribution (orange) is shifted toward smaller cosine distances
relative to the first-pass distribution (blue), with mean
$\bar{\Delta}$ ranging from $-0.113$ (GPT-4o) to $-0.178$
(DeepSeek-R1). This indicates that second-pass responses
systematically move closer to the model's own task prior,
consistent with distributional attraction rather than targeted
error correction. Notably, DeepSeek-R1---the strongest model in
our evaluation and the one exhibiting the largest apparent
reflection gains on MalAlgoQA---shows the most pronounced drift
toward its prior, suggesting that even its improvements may
reflect regression to a strong default rather than genuine
error-specific revision.

In contrast, human annotators exhibit a slight \emph{positive}
mean drift ($\bar{\Delta} = +0.086$), with the first-pass and
reflection distributions largely overlapping. Human revisions do
not systematically converge toward a shared prior; instead, they
remain localized and item-specific, consistent with selective
error correction rather than default-mode re-sampling.

\newpage

\section{Diminishing Returns Across Iterations}
\label{app:diminishing}

\begin{figure*}[t]
\centering
\includegraphics[width=0.95\linewidth]{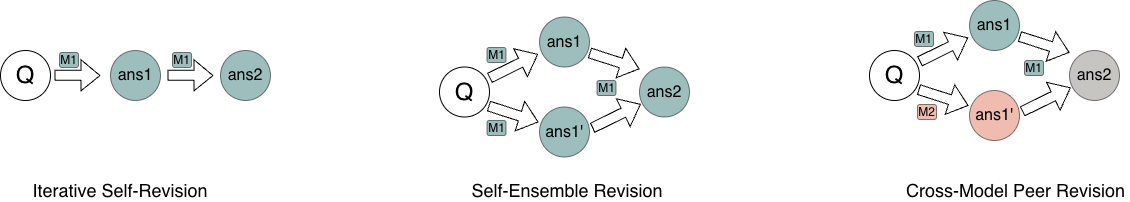}
\caption{Three multi-iteration revision strategies evaluated in this
appendix. \textbf{Left:} Iterative self-revision, where model M1
repeatedly revises its own prior output.
\textbf{Center:} Self-ensemble revision, where M1 generates two
independent first-pass responses and revises by conditioning on both.
\textbf{Right:} Cross-model peer revision, where M1 revises after
seeing a first-pass response from a different model M2.}
\label{fig:revision-strategies}
\end{figure*}

The main text establishes that a single reflection step yields
near-zero or negative information gain. A natural follow-up question
is whether \emph{repeated} revision can overcome this ceiling by
accumulating small gains across iterations. We test this by measuring
per-iteration information gain across up to five revision passes on
MalAlgoQA under three strategies:
\emph{iterative self-revision}, in which the same model revises its
own output repeatedly;
\emph{self-ensemble revision}, in which independent first-pass
samples from the same model serve as prior responses; and
\emph{cross-model peer revision}, in which a different model's
output serves as the prior response.
These strategies are illustrated in Figure~\ref{fig:revision-strategies}.

\begin{figure*}[t]
\centering
\begin{minipage}{0.48\linewidth}
  \centering
  \includegraphics[width=\linewidth]{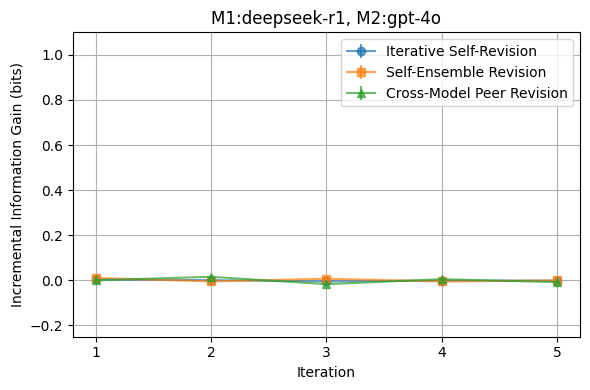}
  \centerline{\small (a) Primary: DeepSeek-R1; auxiliary: GPT-4o.}
\end{minipage}
\hfill
\begin{minipage}{0.48\linewidth}
  \centering
  \includegraphics[width=\linewidth]{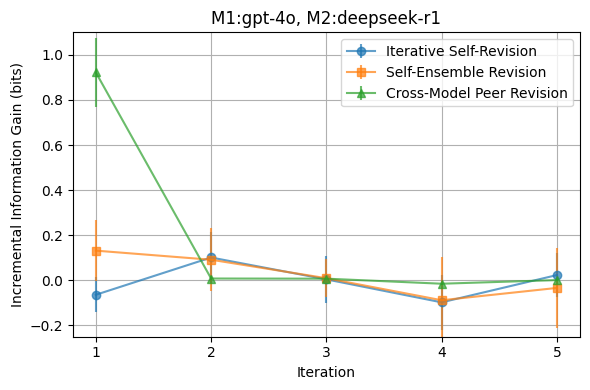}
  \centerline{\small (b) Primary: GPT-4o; auxiliary: DeepSeek-R1.}
\end{minipage}
\caption{Incremental, label-anchored information gain per revision
iteration on MalAlgoQA for iterative self-revision, self-ensemble
revision, and cross-model peer revision.}
\label{fig:ig-main}
\end{figure*}

Figure~\ref{fig:ig-main} reports incremental information gain per
iteration under both primary--auxiliary model assignments, obtained by
swapping the roles of GPT-4o and DeepSeek-R1.

Across all strategies and both assignments, incremental information
gain is largely confined to a single revision step and collapses to
near zero thereafter. Iterative self-revision saturates immediately.
Self-ensemble and cross-model revision occasionally produce larger
first-step updates with high variance but no persistence across
iterations. A transient gain appears only when the auxiliary model is
stronger than the primary model, and only at the first step; reversing
the assignment eliminates the effect, yielding near-zero gain from the
outset. Despite the two models making different errors on
approximately $30\%$ of items, such error diversity does not translate
into sustained improvements under iterative revision.

These results confirm that the information ceiling is structural:
computational resources spent on iterative self-revision yield
diminishing returns after the first pass and are better allocated to
independent sampling or external verification.

\section{Additional Model Results}
\label{app:additional-models}

To verify that the reflection failures documented in the main text are
not confined to specific model versions, we evaluate two recent GPT-5
series models under the same HRF two-pass protocol. Identical prompts
and evaluation procedures are used throughout.

Table~\ref{tab:additional-models} reports mean accuracy and standard
deviation over $K{=}5$ repeated runs on MalAlgoQA.

Across both models, second-pass revision does not produce systematic
improvements. Instead, we observe small but consistent performance
decreases, along with increased variance in the second pass. These
results confirm that the conditioned re-generation behavior documented
in the main text persists in more recent and stronger reasoning
systems.

\begin{table}[h]
\centering
\small
\begin{tabular}{lccc}
\toprule
Model & 1st Pass & 2nd Pass & $\Delta$ \\
\midrule
GPT-5-mini    & 96.78 {\scriptsize$\pm$ 0.27}
              & 96.36 {\scriptsize$\pm$ 0.50}
              & \textcolor{red}{$-$0.42} \\
GPT-5.2       & 84.21 {\scriptsize$\pm$ 0.35}
              & 83.14 {\scriptsize$\pm$ 0.75}
              & \textcolor{red}{$-$1.07} \\
\bottomrule
\end{tabular}
\caption{MalAlgoQA accuracy (\%) for GPT-5 series models under the
HRF self-revision protocol. $\Delta$ is the change from first to
second pass in percentage points.}
\label{tab:additional-models}
\end{table}

\end{document}